# Random forest model identifies serve strength as a key predictor of tennis match outcome

**Zijian Gao**[1], **Amanda Kowalczyk**[2]
[1]Darlington School, Rome, GA
ggao@darlingtonschool.org
[2]Carnegie Mellon University-University of Pittsburgh
PhD Program in Computational Biology, Pittsburgh, PA
kowaae22@pitt.edu

**Abstract.** Tennis is a popular sport worldwide, boasting millions of fans and numerous national and international tournaments. Like many sports, tennis has benefitted from the popularity of rigorous record-keeping of game and player information, as well as the growth of machine learning methods for use in sports analytics. Of particular interest to bettors and betting companies alike is potential use of sports records to predict tennis match outcomes prior to match start. We compiled, cleaned, and used the largest database of tennis match information to date to predict match outcome using fairly simple machine learning methods. Using such methods allows for rapid fit and prediction times to readily incorporate new data and make real-time predictions. We were able to predict match outcomes with upwards of 80% accuracy, much greater than predictions using betting odds alone, and identify serve strength as a key predictor of match outcome. By combining prediction accuracies from three models, we were able to nearly recreate a probability distribution based on average betting odds from betting companies, which indicates that betting companies are using similar information to assign odds to matches. These results demonstrate the capability of relatively simple machine learning models to quite accurately predict tennis match outcomes.

# Introduction

Tennis has been one of the most popular sports in the world for decades. Professional tennis tournaments, after years of advancement and improvement, have developed into the ATP Masters, ATP Cup, and the Grand Slams, which include the US Open, French Open, Wimbledon Open, and Australian Open [1]. In tandem, sports analytics has thrived in the 21$^{st}$ century, with applications in various sports, such as predicting passes in soccer, optimizing player selection in baseball and hockey, assessing football plays, and broadly using massive sports datasets to guide team training and strategy [2]. Likewise, sports betting has become a popular pastime, with a global multi-billion-dollar market. Betting is handled through betting companies and bookkeepers who assign odds to sports outcomes. Bettors win the bet if they bet on the correct outcome, and the prize is determined by the odds. For example, if a person bets a dollar on 2:1 odds (decimal odds 1.5), the person will win two dollars if the outcome is in his or her favor.

As internet-betting has become more popular around the world, the accuracy of betting odds, set by the bookmakers, has become more crucial. The most popular betting frameworks fall into two categories: parimutuel betting, where odds are calculated after bets are placed based on the amount bettors bet, and fixed betting, in which odds are published before the match starts. The latter, fixed betting, is used for tennis betting [3]. Therefore, prior to the match start, the implied probability of a player winning a match can be calculated by a linear transformation of
$$P = \frac{1}{ODDS}$$
in which ODDS represents the decimal odds of a player winning and P represents their probability of winning.

To attempt to predict the result of a tennis game based on the match information and the player's data, bookmakers from betting companies use mathematical machine learning models based on various characteristics of players, tournaments, and matches to assign betting odds. Previous studies have attempted to recapitulate the methods used by bookmakers to accurately predict tennis match outcomes. One study used multiple multilayer perceptrons, including StatEnv, AdvancedStatEnv, and TimeSeries Models, to predict 2007 and 2008 Grand Slams' match results and match length with upwards of 70% accuracy [4]. They excluded data that was more than two years prior to the match date and included environmental data such as court surfaces [4]. Another study used Markov chain models to predict the 2003 Australian Open results and analyzed how to bet [5]. Some studies focus on more in-depth characteristics of specific matches, such as research that uesd Markov chain models to find out the correlation between match data and duration by exampling the match between Roddick and El Aynaoui played at the 2003 Australian Open [6].

We go a step further than previous work by compiling the largest dataset to date of tennis match statistics to use to train our machine learning models. We include a wide variety of features to capture information about physical, psychological, court-related, and match-related variables that may help predict match outcome compiled and processed from ATP data from 2000 to 2016. Environmental variables such as different court surfaces, including clay, grass, and hard court, can make a difference on player's performance [6]. For example, Rafael Nadal has won 12 titles on clay courts in the French Open, which is two thirds of all grand slam titles he has won [7]. Players' physical statistics, such as height and age, are included as factors since they may determine match outcome due to physical advantages of particular players [8]. We also include psychological variables, such as the percentage of break points saved, round number, and previous results playing against the same opponent, since psychological momentum has previously been suggested to directly affect sports success and failure [9]. Previous studies also discovered that first serve is a top significant factors to predict tennis match outcome, and this study further includes the variable of percent accuracy for first serves as well as success in second serves [10]. In addition, more typical measures of player proficiency, such as player rank, are also included [11]. To predict match outcomes, we employ relatively simple models (support vector machine, logistic regression, and random forest classifier) that allow easy interpretability and fast training times that facilitate inclusion of additional data as it becomes available. In doing so, we were able to achieve excellent match outcome prediction accuracy, as well as identify key components to predict match outcomes.

## Methods

The final dataset used in this study was compiled based on numerous other datasets from the ATP, including the ATP World Tour which is comprised of the ATP World Tour Masters 1000, ATP World Tour 500 series, ATP World Tour 250 series, and the ATP Challenger Tour. By merging the datasets from 2000 to 2016 provided by atptennis.com (the official website for ATP), it was possible to compile all the data from every match from each major ATP tournament, including environmental data, general match information, match results, and betting odds from major betting companies. We also used information from the Match Charting Project uploaded on GitHub by JeffSackmann, which included the betting odds of major bookmakers such as Betting365 [12]. After taking the average of all the betting odds from various companies for each match and combining the two datasets via the match date and players' names, we're able to get a big dataset with both match information and betting odds.

To clean the data, we filled in missing data values using the median value for each feature to allow us to use the full dataset while assigning values that were unaffected by potentially skewed data and would not have a detrimental impact on model fit [13]. Some variables were combined to represent various characteristics of player performance as shown in Table 1. For example, we used ace divided by double faults instead of using ace or double faults alone because the value represents how aggressive their play style is and how accurately their serves are. By combing the values, we minimize multiple collinearity problems among our variables while still using as much information as possible. We also created new variables to represent players' records over a set period of time to create statistics that represent up-to-date player performance. For example, the number of games played in past 12 months could show whether the player is active or injured for the past year, as well as how seriously they are competing and attending tournaments.

The full dataset has 49,188 entries where each entry represents one player in one match. The full dataset, split into test and train portions, was used to calculate prediction accuracy and perform feature selection. To compare our model predictions to betting odds, models were fitted to the subset of data that do not have betting odds information available (29,238 entries) and tested on the entries that do have betting odds information available (19,880 entries).

We used three machine learning methods: support vector machine with a radial basis function kernel, random forest classification, and logistic regression, to attempt to predict match outcomes based on the variables listed in table one. Model accuracy was assessed based on test accuracy using the random train/test split and 10-fold cross validation. We also compared our probabilities to probabilities calculating from betting odds using the following formula:

$$Score = w * (p - 0.5)$$

where p is the probability of winning according to either betting odds or probabilities from machine learning predictions and w is an indicator variable representing whether a player won or lost the match (w=1 indicates that the player won and w=-1 indicates that the player lost). A high score (up to 0.5) indicates that a player was predicted to win a match with high certainty and the player did win the match, or that a player was predicted to lose a match with high certainty and the player did lose. Alternatively, a low score (down to -0.5) indicates that a player was predicted to win a match with high certainty and the player lost the match, or that a player was predicted to lose a match with high certainty but actually won. Scores close to zero indicate low prediction confidence, with positive values indicating correct low-confidence predictions and negative values indicating incorrect low-confidence predictions.

**Table 1.** Features compiled per tennis player, per match.

| Variable | Code | Category | Calculation | Rationale |
|---|---|---|---|---|
| *Height* | w_height | Physical characteristic | Known numeric data | Height increases serving speed |
| *Age* | w_age | Physical characteristic | Known numeric data | Skill increases with age, athleticism decreases with age |
| *Rank Points* | w_rank_points | Record | Based on previous match results for the past year | Accumulated by winning matches, better record indicates better player |
| *Court Surface* | surface* | Court/tournament information | Known categorical data | Some players are better at particular courts |
| *Percentage of Ace over Double Faults* | AceVsDf | Serve | Ace/double faults over past 12 months | Higher value means higher serving speed and accuracy |
| *Previous percentage of games won* | PastPer | Record | Games won/games played over the past 12 months | Better record indicates better player or a win streak |
| *Numbers of championship* | Champ | Record | Known numeric data, lifetime | Better record indicates better player |
| *Games won before in the same round* | WinRound | Record | Known numeric data | Some players play better in final rounds |
| *Games played in past 12 months* | GamesPlayed | Record | Known numeric data | Health condition |
| *Percentage of games won in the same tournament* | TourPer | Record | Games won in the tournament/games played in the tournament | Some players play better in certain tournaments |
| *Percentage of games won against player with same handedness as the current opponent* | HandPer | Physical characteristic | Games won against player with same handedness/games played against player with same handedness | Lefthanders tend to play better against righthanders |
| *Percentage of games won on the same type of surface* | SurfacePer | Court/tournament information | Games won on current surface/games played on current surface | Some players play better on certain surfaces |
| *Percentage of games won against the same opponent* | OpponPer | Mental Strength | Games won against current opponent/games played with current opponent | Some players play better against certain opponents |
| *Percentage of making the first serve* | FirstIn1stServe | Serve | First serves that go in/first serves | First serves' accuracy |
| *Percentage of making first serve and winning* | FirstWonFirstIn | Serve | First serves that help the player win the point/first serves that go in | First serves' power |
| *Percentage of making second serve and winning* | SecondWonSecondIn | Serve | Second serves that help the player win the point/second serves that go in | Second serves' power |
| *Percentage of break points saved* | BpSBpF | Mental Strength | Break points saved/break points faced | Mental strength |
| *Current round number* | round* | Court/tournament information | Known categorical data | Some players play better on particular rounds |

# Results

Optimal features were selected by systematically testing model accuracy on subsets of features listed in Table 1. Both fitted support vector machine and logistic regression models performed best using the full set of features, but the random forest model was more sensitive to choice of input parameters (Figure 1). In particular, removing either the "First Won First In" and "Second Won Second In" variables alone caused a large drop in test accuracy (Figure 1), and when features were added sequentially, adding those two variables caused the largest jump in increased accuracy (Figure 2), which indicated that these variables are important to model fit.

When further optimizing feature selection for the random forest model, some features, when included in the model, were found to actually decrease model test accuracy (Figure 3). We removed features that decreased accuracy and obtained a 76.23% 10-fold cross validation accuracy compared to 73.85% accuracy when using all features. We then, one-by-one, returned the previously removed features to the model, each of which increased the model accuracy when added singly (Figure 4). Using those accuracies, we then cumulatively added features in an order based on how much they increased test accuracy when added singly – features that improved accuracy the most were added first. Adding the "RoundSF" variable increased the accuracy the most, to 80.68%, but adding any additional variables decreased accuracy, so our final feature set for the random forest model included "w_height", "w_age", "AceVsDf", "Champ", "GamesPlayed", "FirstIn1stServe", "FirstWonFirstIn", "SecondWonSecondIn", "roundR128", and "roundRR".

Final 10-fold cross validation accuracies for all models and predictions based on betting odds are shown in Table 2. Overall, the random forest model showed the highest prediction accuracy at 83.18% compared to accuracy of 69.04% when using predictions based on betting odds probabilities. However, betting odds showed the highest score of 2059.66, followed by the support vector machine model score of 1750.91. The discrepancy between accuracy and score can be explained by the distribution of scores as shown in Figure 5. While most predictions from the random forest model are correct, these predictions are made with low confidence, so they have lower scores. The same is true for the logistic regression model, although it also demonstrates a lower accuracy. Both the support vector machine model and the probabilities from betting odds show a more extreme shift toward higher scores, which gives them a higher overall score even though they show lower prediction accuracy.

## Discussion

By using machine learning models, performing in-depth feature selection, and compiling the largest tennis statistics data set available, we were able to accurately predict tennis match outcome with upwards of 80% accuracy. Our accuracy exceeded prediction accuracy based on betting odds, which indicates that we were able to predict match outcome with unprecedented probability. We also identify serve strength, as represented by the proportion of first and second serves for which a player won a point, as a key predictor of match outcome.

The importance of serve strength to predicting match outcome is unsurprising given the importance of athleticism and strength to overall tennis performance. Serve strength, particularly in the first serve, sets the stage as an offensive attack for the rest of the game, and since many points are scored through a powerful serve, serve strength is also a key scoring method throughout a match [14]. A variety of factors can impact serve strength, including the angle of trunk rotation [15], knee bend angle and extension [16], right shoulder angle [17], and a successful serve requires success in all eight stage of a serve (start, release, loading, cocking, acceleration, contact, deceleration, and finish) [18]. A fruitful area of endeavor to improve match outcome prediction may be to measure these statistics for each player and incorporate it as a model parameter.

In addition to having better serves, players who serve better may also be in largely better physical condition, which in turn improves their overall game performance. Note that although serve strength is a key predictor in our random forest model, this does not necessarily imply that a player will win more games simply by improving his serve. The information we use about serve strength likely encapsulates a plethora of information about overall player proficiency rather than indicating a specific area of focus for players who want to improve their tennis skills.

Interestingly, features one may expect to indicate player skill, such as rank and winning record, had relatively little impact on prediction accuracy. Similarly, other external factors such as tournament round and court surface, as well as potential psychological factors such as proportion of break points saved and record against the current opponent, had little impact on overall prediction accuracy. These findings may seem in direct opposition to previous modeling results, which found that player performance was significantly impacted by court surface type and round number when match data were analyzed on a point-by-point basis [19]. However, we argue that match-specific performance may not necessarily be indicative of overall performance, and player success over larger timescales is based more on player skill than on external factors.

Physical characteristics such as player age and height did improve model accuracy, although not as much as serve strength. This is consistent with previous work that found a significant positive correlation between player height and number of aces (an unreturned successful serve, winning the point) [20]. Surprisingly, although serve strength was a key predictor, serve accuracy seemed less important. Features such as ace versus double faults and proportion of first serves that go in, which represent serve accuracy, although they did improve model prediction accuracy, they did not impact accuracy as much as serve strength, perhaps because they do not as effectively represent overall player physicality.

    Although we have no information about how tennis betting odds were calculated, we do find it surprising that our model is better able to predict match outcome than probabilities based on the betting odds. We suspect this is due either to technical reasons, such as a smaller data set or a worse set of training regions used to assign betting odds, or, more likely, due to financial reasons associated with balancing betting payouts. For example, if a very famous, highly-ranked player is predicted to lose a match, the betting company may still assign him good betting odds to win money from bettors who bet on him to win, and vice versa for unknown, low-ranked players. Bookmaker bias, or odds that are too low for favorites and too high for long-shots, have been found in numerous betting markets [21], and are likely present in tennis betting odds assignment as well. This bias can exist for a variety of reasons, including establishing betting equilibrium based on risk-attitudes [22], mental factors underlying betting decisions [23], and the existence in insider bettors [24]. We also note that probabilities from betting odds better demonstrate high confidence correct predictions and low confidence incorrect predictions, which our model, despite high prediction accuracy, fails to accomplish.

    Overall, we have constructed a model that predicts tennis match outcomes with high accuracy and identified serve strength as a key predictor. We were able to do so using a random forest classifier that highlights the importance and relevance of simple models in the age of deep learning. These findings may be used to predict match outcomes and to guide similar future endeavors to investigate tennis match results.

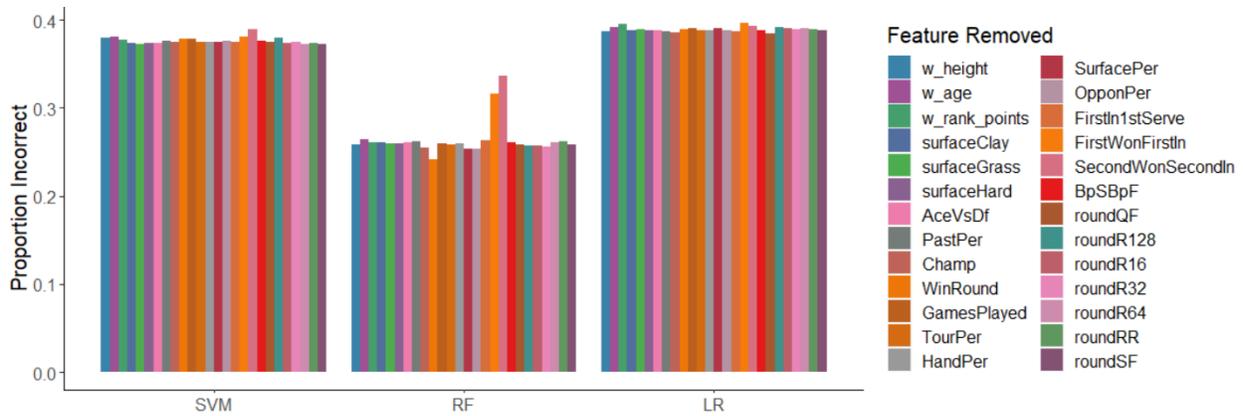

**Figure 1.** Model test prediction accuracy when removing individual features. Individual feature removal has little impact on support vector machine and logistic regression test accuracy, but "FirstWonFirstIn" and "SecondWonSecondIn" markedly decrease accuracy of random forest test predictions when removed.

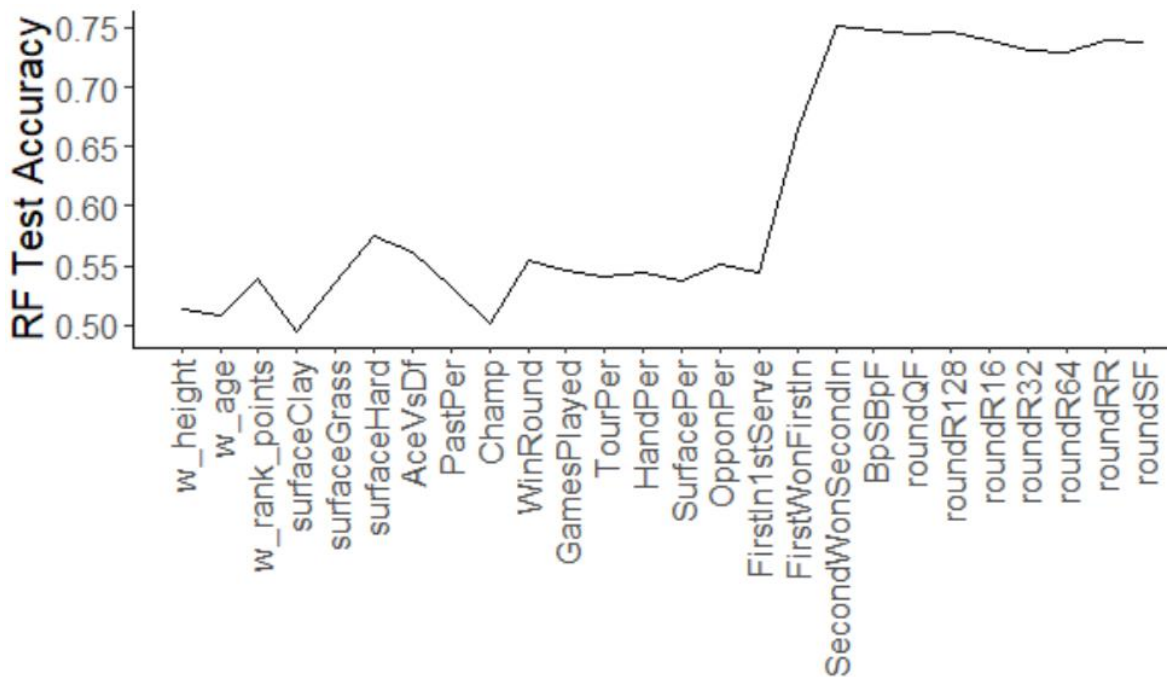

**Figure 2.** When adding features to the random forest model one-by-one, "FirstWonFirstIn" and "SecondWonSecondIn" show the most dramatic accuracy increases.

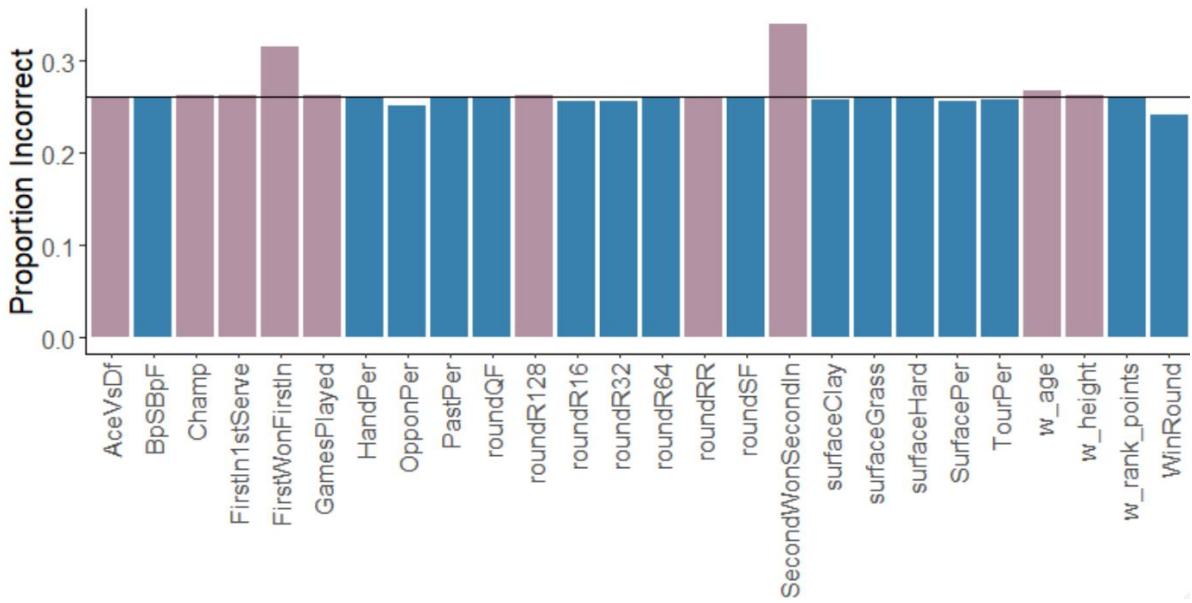

**Figure 3.** For the random forest model, 10-fold cross validation accuracy when individual features are removed. The black line indicates the 10-fold cross validation accuracy using all features. Features depicted in pink (lighter color) decrease accuracy when removed and were therefore included as important features. All other features increased accuracy when removed.

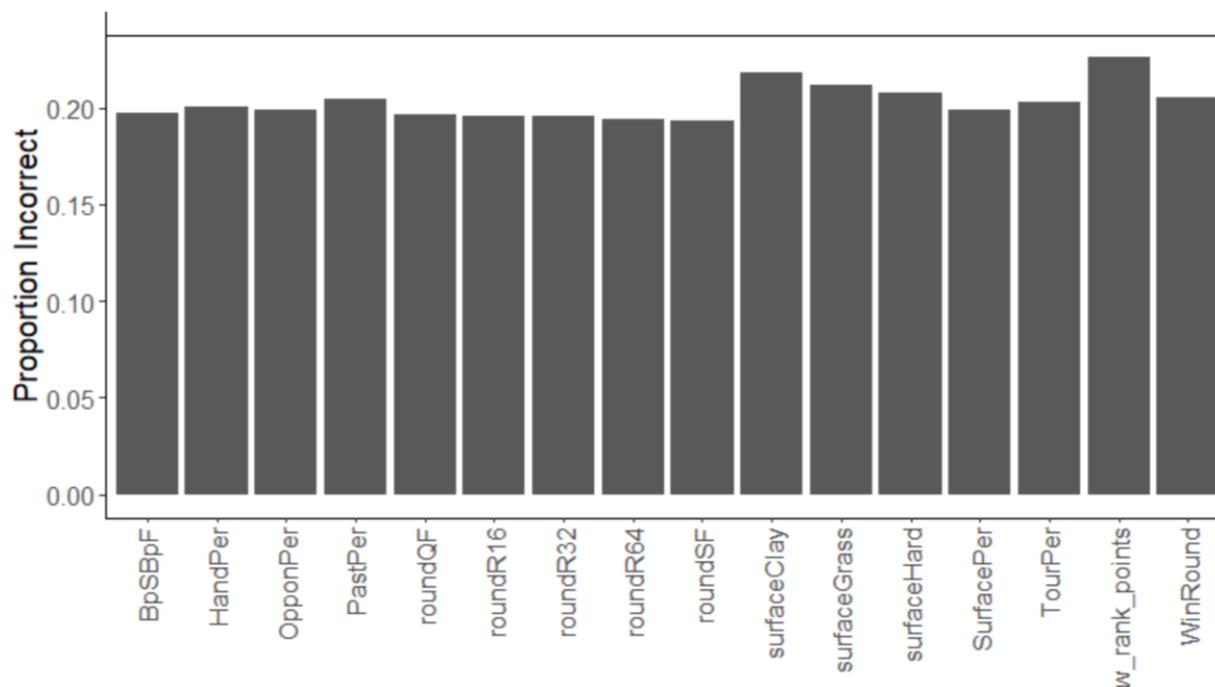

**Figure 4.** Random forest 10-fold cross validation accuracy when previously-removed features are added to the feature set. The black line indicates the 10-fold cross validation accuracy using all previously-selected features.

**Table 2.** Final 10-fold cross validation accuracy for all models and probabilities from betting odds. Note that betting odds accuracy is only computed on the subset of data that have betting odds available.

| | Final 10-fold Cross Validation Accuracy | | | |
|---|---|---|---|---|
| | Logistic Regression | Random Forest | Support Vector Machine | Betting Odds |
| Percent Correct | 62.06% | 83.18% | 61.60% | 69.04% |
| Score | 834.04 | 884.36 | 1750.91 | 2059.66 |

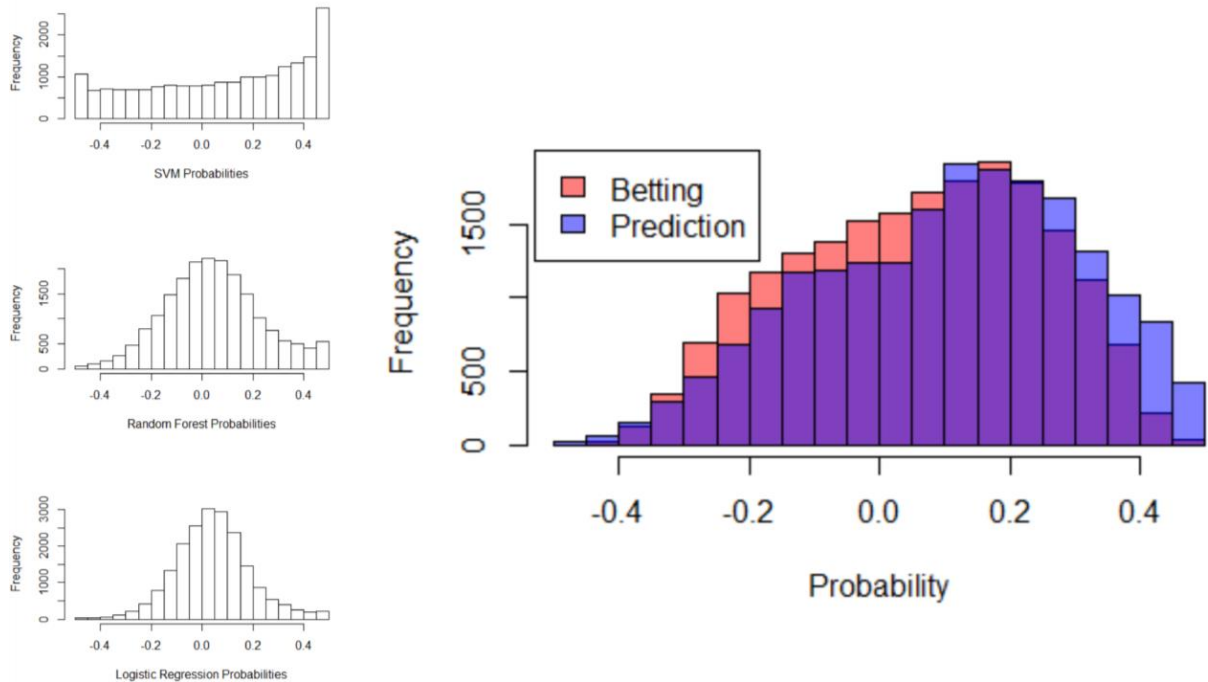

**Figure 5.** Score distribution for all models and average scores from all models alongside betting odds scores. High scores indicate high-confidence correct predictions and low scores indicate high-confidence incorrect predictions. Scores near zero are low-confidence predictions, with positive values indicating correct predictions and negative scores indicating incorrect predictions. All distributions are shifted to the right (indicating more correct predictions than incorrect predictions).